\documentclass[lettersize,journal]{IEEEtran}
\usepackage{amsmath,amsfonts}
\usepackage{algorithmic}
\usepackage{algorithm}
\usepackage{array}
\usepackage[caption=false,font=normalsize,labelfont=sf,textfont=sf]{subfig}
\usepackage{textcomp}
\usepackage{stfloats}
\usepackage{url}
\usepackage{verbatim}
\usepackage{graphicx}
\usepackage{cite}
\usepackage{soul}
\begin{document}

\title{LightZeroNav: Zero-Shot Vision Language Navigation in Continuous Environments Based on Lightweight VLMs
}

\author{Kun Luo$^{\dagger}$, Xiangyu Dong$^{\dagger}$, Xiaoguang Ma$^{*}$, Haoran Zhao, and Yaoming Zhou$^{*}$%
\thanks{$^{\dagger}$Kun Luo and Xiangyu Dong contributed equally to this work.}%
\thanks{$^{*}$Corresponding authors: Xiaoguang Ma and Yaoming Zhou.}%
\thanks{Kun Luo, Xiangyu Dong, and Xiaoguang Ma are with the Foshan Graduate School of Innovation, Northeastern University, Foshan 528311, China (e-mail: lkkun99668@163.com; dxy1999ai@163.com; maxg@mail.neu.edu.cn).}%
\thanks{Xiaoguang Ma is also with the Faculty of Robot Science and Engineering, Northeastern University, Shenyang 110819, China.}%
\thanks{Haoran Zhao and Yaoming Zhou are with the School of Aeronautic Science and Engineering, Beihang University, Beijing 100191, China (e-mail: zhaohaoran@buaa.edu.cn; zhouyaoming@buaa.edu.cn).}%
\thanks{Xiaoguang Ma and Haoran Zhao are also with QingniaoAI, China.}%
}



\maketitle

\begin{abstract}
Although vision-language navigation (VLN) has progressed rapidly, zero-shot VLN in continuous environments (VLN-CE) remains highly challenging when using lightweight vision-language models (VLMs), whose limited reasoning capacity makes long-horizon navigation unreliable. In this paper, we propose \textbf{LightZeroNav} to tackle the three major bottlenecks when using lightweight VLMs in zero-shot VLN-CE,i.e.,information redundancy from multi-source inputs, inaccurate progress estimation caused by noisy textual memory, and task entanglement between action execution and stage transition. Using only RGB observations and a lightweight open-source Qwen3-VL-8B backbone, LightZeroNav achieves  competitive performance with GPT-4o ($\sim$200B) without task-specific training, graph search, or waypoint predictors, demonstrating its effectiveness in zero-shot VLN-CE.
\end{abstract}


\section{Introduction}
\IEEEPARstart{V}{ision} Language Navigation (VLN) requires an embodied agent to follow natural-language instructions in a visually grounded 3D environment \cite{anderson2018vision,savva2019habitat}. VLN in Continuous Environments (VLN-CE) is particularly challenging because the agent must perform fine-grained continuous control under partial observability over long horizons, without privileged graph structure, oracle waypoints, or localization \cite{krantz2020beyond}.

Learning-based VLN has shown that long-horizon navigation benefits not only from strong state representations \cite{wu2024vision,yu2025mossvln,deng2026video}, but also from safe control policies \cite{yue2024safe}, explicit intermediate planning structures \cite{zhang2025flexvln}, and adaptive reasoning mechanisms for efficient inference and environmental generalization \cite{tan2025source}. This also extends to zero-shot VLN-CE. Existing structured approaches improve long-horizon navigation through explicit environmental organization, such as sub-instruction management, value maps, waypoint-oriented scaffolds, or switching criteria \cite{shi2025smartway,long2024instructnav,chen2025constraint}. Meanwhile, recent prompt-centric methods leverage the strong reasoning capability of VLMs by repeatedly feeding observations, navigation history, and language context back into the model for step-wise decision-making \cite{qiao2025open,shi2025fast}. However, most existing zero-shot VLN-CE methods rely heavily on closed-source large-scale VLMs to maintain reasoning performance and execution robustness\cite{long2024instructnav,shi2025fast,shi2025smartway}. In contrast, lightweight open-source VLMs remain relatively underexplored in zero-shot VLN-CE due to their limited reasoning capacity, especially for long-horizon embodied decision-making.

In fact, the limitations of using lightweight VLMs in zero-shot VLN-CE exists at distinct levels. First, \textbf{information redundancy} hurts since decision-making depends on voluminous multi-source inputs, including panoramic multi-view images, complete navigation instructions, and accumulated historical context, which exceed the capacity bottleneck of lightweight VLMs. Redundant task-irrelevant or progress-inconsistent features also disperse limited attention, resulting in a poor signal-to-noise ratio for decision-making \cite{qiao2025open,shi2025fast}.
Second, \textbf{progress estimation inaccuracy} arises because progress estimation of conventioned VLN-CE methods is inferred from textual historical summaries converted from visual observations\cite{shi2025fast}.  This inevitably introduces information loss and may also contain hallucinated or imprecise descriptions. Lightweight VLMs tend to over-rely on these compressed histories due to limited reasoning capacity and lack a strong ability to detect or correct such potential summarization errors. As a result, noisy or incorrect historical cues are treated as reliable signals, leading to unreliable progress estimation and increased semantic drift.
Third, \textbf{task entanglement} occurs when decision identification, action execution, and stage transition are conflated into a single inference step without explicit decoupling. This entanglement enlarges the effective decision space, exceeds the reasoning capacity of lightweight VLMs, and leads to error accumulation and propagation \cite{long2024instructnav,chen2025constraint,shi2025fast,qiao2025open}.

To address these issues, we propose \textbf{LightZeroNav}, a zero-shot framework that formulates VLN-CE as a stage-structured decision process to enhance its reasoning capability with lightweight VLMs. In summary, we makes the following contributions:
\begin{itemize}
 \item We introduce \textbf{Decision-guided Multi-Feature Filtering (DGMF)} to mitigate information redundancy and attention dilution in multi-source visual inputs, so that even lightweight VLMs can maintain high navigation efficiency under redundancy-heavy inputs.
 
\item For progress estimation inaccuracy, we propose \textbf{Image-level State Grounding Representation (ISGR)} for implicit navigation state modeling. ISGR adopts a dual-reference mechanism combining short-term trajectory observations with periodic visual baselines, remarkably improving progress estimation accuracy through image-level comparative reasoning without textual memory maintenance, and reducing hallucination accumulation and semantic drift during long-horizon navigation.

 \item  We design a \textbf{Stage-wise Execution--Transition Decoupling (SETD)} paradigm to alleviate sub-goal entanglement and stage confusion in long-horizon navigation. SETD decomposes instructions into ordered sub-goals and separates inference into an execution stage for local action generation and a transition stage for sub-goal verification and global progress control, ensuring that each inference step focuses on a single decision objective and reducing error propagation across sub-goals.
 \item  Comprehensive tests show that LightZeroNav can achieve competitive zero-shot performance with GPT-4o($\sim$200B) even using lightweight open-source Qwen3-VL-8B backbone with only RGB observations, without task-specific training or fine-tuning, waypoint predictors, or external spatial scaffolds. 
\end{itemize}

\section{Related Work}
\subsection{Zero-shot vision language navigation in continuous environments (VLN-CE)}

VLN is early studied in discrete settings, while VLN-CE extends it to continuous control without predefined topology, oracle actions, or idealized localization \cite{anderson2018vision,krantz2020beyond}. Recent work has explored zero-shot navigation with large language or vision language models in CE. Open-Nav studies open-source LLMs for continuous instruction \cite{qiao2025open}; DreamNav introduces trajectory-level imagination \cite{wang2025dreamnav}; InstructNav emphasizes generic instruction planning with value-map-based execution \cite{long2024instructnav}; Fast-SmartWay largely replaces panoramic observations during step-wise execution and removes waypoint predictors for end-to-end action prediction from frontal views \cite{shi2025fast}. Other methods further use waypoint-oriented scaffolds or value structures to stabilize decisions \cite{shi2025smartway,chen2025constraint}.

Althought these studies confirm the effectiveness of using foundation models in VLN-CE, existing zero-shot VLN-CE methods based on VLMs still suffer from three major bottlenecks: information redundancy, progress estimation inaccuracy, and task entanglement. 

\subsection{Information Redundancy}

Input feature management in zero-shot VLN-CE involves a trade-off between information richness and model capacity. Panoramic or multi-view pipelines improve scene coverage and support downstream action selection \cite{krantz2021waypoint,qiao2025open,shi2025smartway}, but they also increase visual redundancy, which can easily exceed the limited reasoning capacity of lightweight VLMs. Fast-SmartWay reduces perceptual cost by replacing panoramas with forward-facing triplet views and relying on step-wise reasoning over accumulated context \cite{shi2025fast}. More broadly, many recent methods follow a \emph{full-input decision} style, where multiple views, history, and auxiliary signals are fed to the model together, often exceeding the capacity bottleneck of lightweight VLMs and dispersing their limited attention across task‑irrelevant or progress‑inconsistent features \cite{qiao2025open,chen2024mapgpt,shi2025smartway,shi2025fast,chen2025constraint}. Moreover, these methods lack selective filtering of decision-relevant evidence, leading to extremely low decision signal-to-noise ratios for lightweight VLMs.

\subsection{Progress Estimation Inaccuracy}

Memory has long been used in embodied navigation through observation histories and retrieval context \cite{georgakis2022cross,wang2023gridmm}. Recent VLN-CE methods further exploit structured priors such as value maps to stabilize zero-shot policies \cite{long2024instructnav,chen2025constraint}. However, these approaches either rely on explicit structured representations or depend on VLM-generated textual summaries to track progress over time.Such text-based progress modeling is inherently susceptible to information compression and ambiguity, which can introduce hallucinated or inconsistent state descriptions, especially under lightweight VLM settings with limited reasoning capacity.

In contrast, this paper estimates navigation progress via image-level contrastive reasoning between short-term observations and a visual baseline, avoiding reliance on textual memory and remarkably mitigating error accumulation in long-horizon navigation.

\subsection{Task Entanglement}

Structured inference has improved reliability in LLM agents. ReAct interleaves reasoning and acting \cite{yao2022react}. Plan-and-Solve emphasizes decomposition \cite{wang2023plan}. Related ideas have also been extended to embodied settings, such as Voyager \cite{wang2023voyager}. In navigation, prior work~\cite{long2024discuss} introduces multi-expert discussion before action selection.

However, existing methods still conflate subgoal identification, action execution, and stage transition into a single inference step\cite{shi2025fast}. This leads to a combinatorial explosion of the decision space, easily exceeding the limited reasoning capacity of lightweight VLMs and causes error accumulation across subgoals.

\section{Method}
\label{sec:method}

\begin{figure*}[t]
    \centering
    \includegraphics[width=1\linewidth]{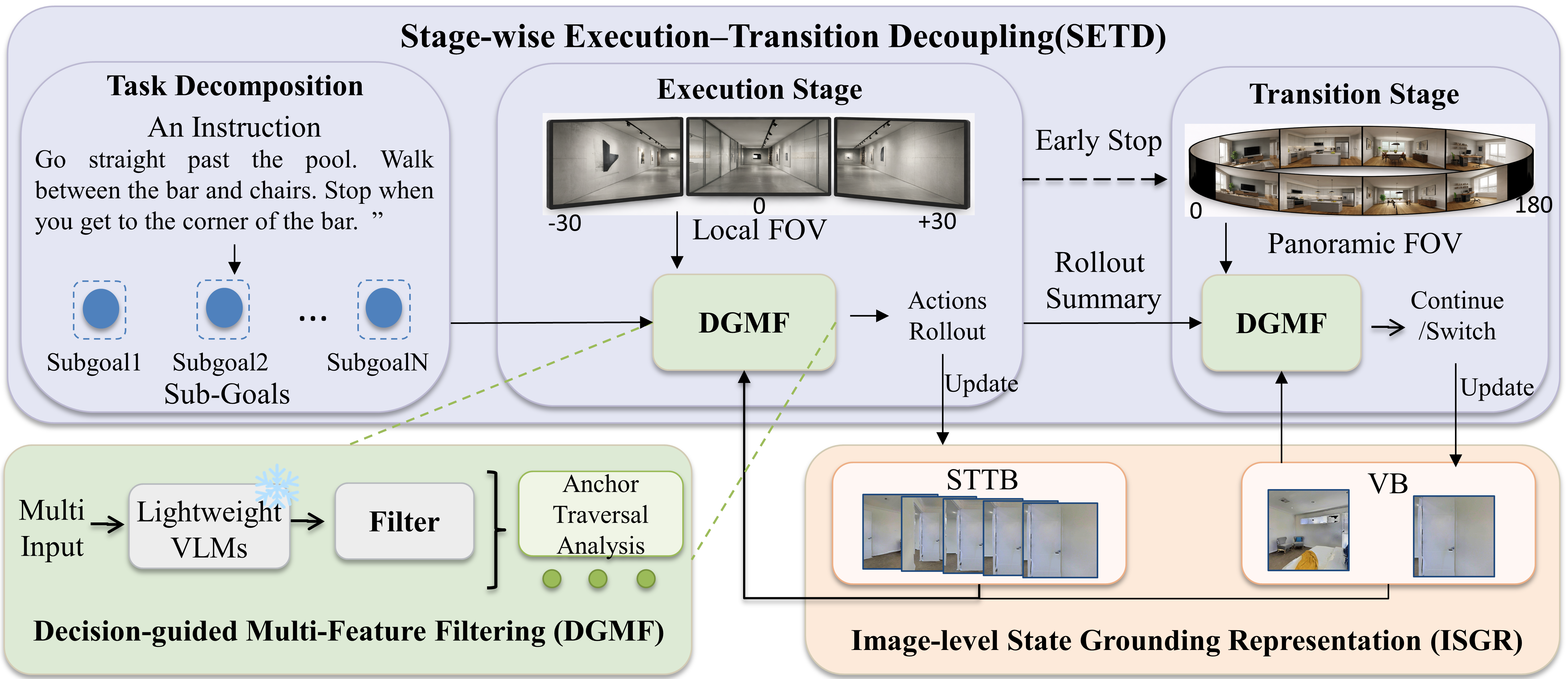}
    \caption{
Overview of the proposed LightZeroNav framework, which enables zero-shot continuous navigation through collaborative optimization across information redundancy, progress estimation inaccuracy, and task entanglement. At the information redundancy level, \emph{Decision-guided Multi-Feature Filtering (DGMF)} filters multi-source observations and trajectory cues according to the current decision objective, retaining only decision-relevant evidence for efficient reasoning. At the progress estimation inaccuracy level, \emph{Image-level State Grounding Representation (ISGR)} combines short-term trajectory observations (STTB) with a visual baseline (VB) to enable implicit progress estimation and positional grounding via image-level comparative reasoning. At the task entanglement level, \emph{Stage-wise Execution--Transition Decoupling (SETD)} decomposes long-horizon instructions into ordered sub-goals and separates inference into an execution stage for local action rollout under local FOV and a transition stage for sub-goal verification and global progress control under panoramic FOV, mitigating sub-goal coupling and decision ambiguity for lightweight VLMs.
    }
    \label{fig:overview}
\end{figure*}

LightZeroNav is built upon a multi-level enhancement framework that progressively reduces inference burden for lightweight VLMs by addressing information redundancy via DGMF, progress estimation inaccuracy via ISGR, and task entanglement via SETD. Figure~\ref{fig:overview} overviews the LightZeroNav framework, and Algorithm~\ref{alg:lightzeronav} summarizes the inference procedure.

\begin{algorithm}[t]
\caption{Inference Procedure of LightZeroNav}
\label{alg:lightzeronav}
\small
\begin{algorithmic}[1]

\STATE Decompose instruction $x$ into ordered sub-goals $G=[g_1,\dots,g_K]$
\STATE Initialize active sub-goal index $k \gets 1$
\STATE Initialize action history $A_{1:T} \gets \emptyset$
\STATE Initialize short-term trajectory observations $\mathrm{STTB} \gets \emptyset$
\STATE Initialize visual baseline $\mathrm{VB}$ from the initial observation

\WHILE{$k \leq K$}

    \STATE Initialize rollout buffer $\mathcal{R} \gets \emptyset$

    \FOR{$h = 1$ to $H$}

        \STATE Extract execution-stage evidence $\mathbf{e}_h^{\mathrm{exec}}$
        
        \STATE Predict action
        $
        a_h \sim
        \pi_{\mathrm{exec}}
        (g_k,
        \mathbf{e}_h^{\mathrm{exec}},
        \mathrm{STTB},
        \mathrm{VB})
        $

        \IF{$k = K$ \textbf{and} $a_h$ is \texttt{stop}}
            \STATE \textbf{break}
        \ENDIF

        \STATE Execute $a_h$

        \STATE Update action history $A_{1:T}$

        \STATE Update short-term trajectory observations $\mathrm{STTB}$

        \STATE Append current observation to rollout buffer $\mathcal{R}$

    \ENDFOR

    \STATE Generate rollout summary
    $
    \Sigma \gets \mathrm{Summarize}(\mathcal{R})
    $

    \STATE Extract transition-stage evidence
    $
    \mathbf{e}^{\mathrm{trans}}
    $

    \STATE Predict sub-goal transition decision
    $
    u \sim
    \pi_{\mathrm{trans}}
    (g_k,
    \mathbf{e}^{\mathrm{trans}},
    \mathrm{VB},
    \Sigma)
    $

    \IF{$u = \mathrm{switch}$}
        \STATE Update visual baseline $\mathrm{VB}$
 \STATE Reset $\mathrm{STTB}$
        \STATE $k \gets k + 1$
    \ENDIF

\ENDWHILE

\end{algorithmic}
\end{algorithm}

\subsection{Decision-guided Multi-Feature Filtering (DGMF)}
\label{sec:tgee}
\begin{figure*}[t]
    \centering
    \includegraphics[width=1\linewidth]{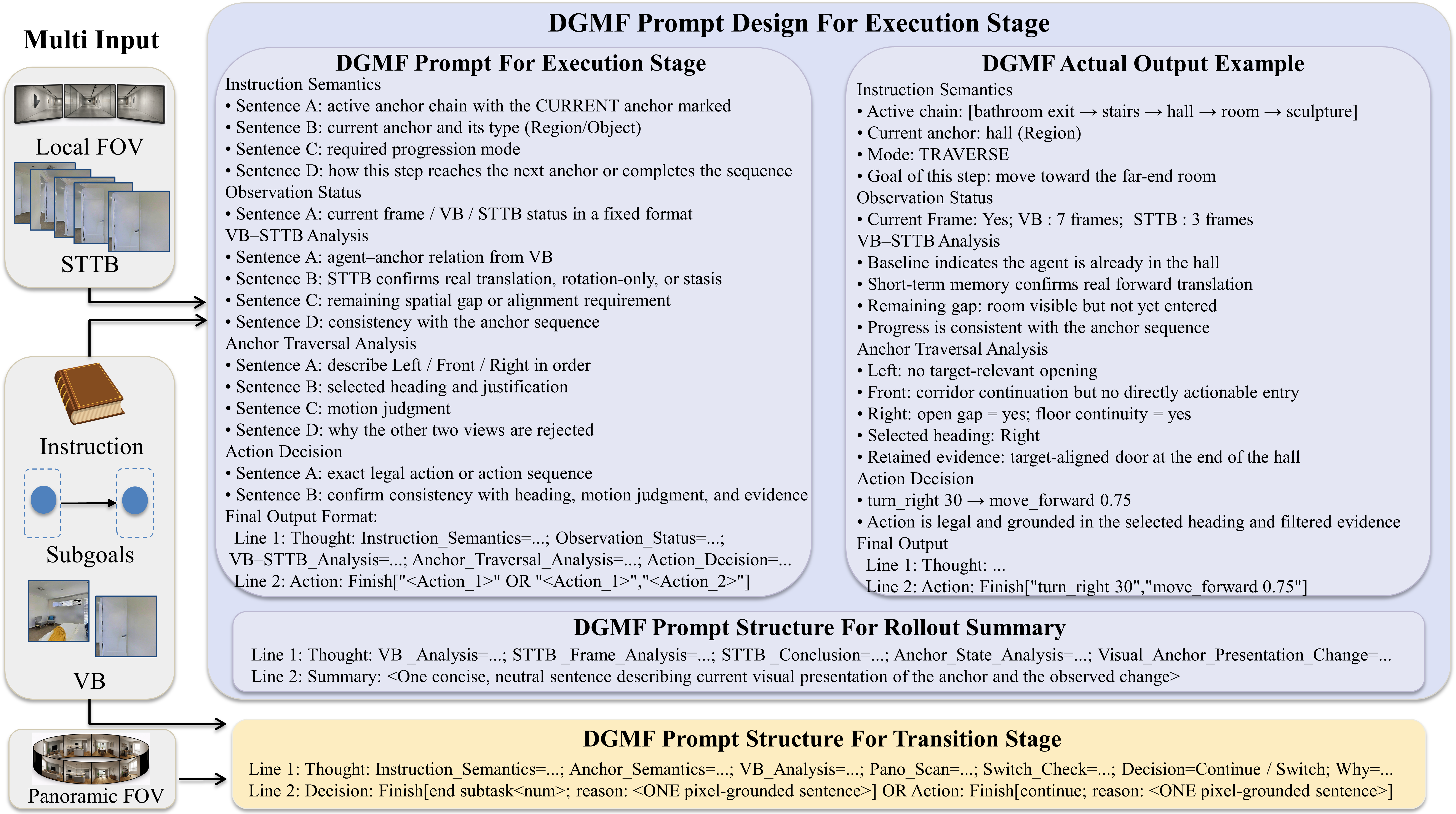}
    \caption{
The prompt-based implementation of the Decision-guided Multi-Feature Filtering(DGMF). It presents three structured prompt designs for execution, rollout summary, and transition, and an actual execution-stage output example showing how decision-relevant evidence is organized into structured reasoning traces.
    }
    \label{fig:4}
\end{figure*}

Long-horizon VLN-CE produces abundant multi-view observations, many of which are irrelevant to the current navigation objective and introduce severe information redundancy that distracts lightweight VLMs from decision-critical cues. When directly processing raw multi-view inputs, lightweight VLMs suffer from information redundancy, making it difficult to maintain attention on subgoal-relevant visual evidence. To address this issue, we introduce Decision-guided Multi-Feature Filtering (DGMF) to explicitly reduce information redundancy and mitigate attention dilution in multi-source visual inputs. DGMF filters multi-view observations and trajectory cues conditioned on the current decision objective, removing redundant information while preserving decision-relevant evidence. This way, redundant multi-source visual features are transformed into compact evidence representations that are directly aligned with the current navigation decision.

Using plug-and-play prompt templates (Fig.~\ref{fig:4}), DGMF guides the VLM to organize decision-relevant evidence into predefined reasoning fields (e.g., instruction semantics, VB-STTB analysis, and anchor traversal) before generating the final action decision. This prompt structure encourages lightweight VLMs to focus their limited reasoning capacity on decision-critical evidence without any parameter updates.  Formally, the filtering process is defined as:

\begin{equation}
\mathbf{e}_t = \mathrm{Filter}\!\left(
g_{k_t}, \phi_t, O_t, M_t, \Sigma_t
\right),
\label{eq:tgee}
\end{equation}

where $M_t = (\mathrm{STTB}_t, \mathrm{VB}_t)$ denotes the dual‑reference state maintained by ISGR (Sec.~\ref{sec:isgr}), $\Sigma_t$ is the rollout summary, and $\phi_t$ indicates the current decision stage (provided by SETD, Sec.~\ref{sec:setd}).

Despite being implemented purely via prompting, DGMF effectively reduces information redundancy and visual distractions, enabling lightweight VLMs to focus on decision-relevant evidence in long-horizon navigation.

\subsection{Image-level State Grounding Representation (ISGR)}
\label{sec:isgr}


Long-horizon VLN-CE requires a stable state representation that enables lightweight VLMs to maintain reliable progress awareness across extended navigation trajectories. Existing zero-shot methods often rely on accumulated textual summaries converted from visual observations to track navigation progress. However, such image-to-text historical summaries inevitably suffer from information loss and hallucinated or imprecise descriptions, and lightweight VLMs tend to over-rely on these compressed histories due to limited reasoning capacity, leading to unreliable progress estimation.

To address this issue, we propose the \textbf{Image-level State Grounding Representation (ISGR)}, a dual-reference grounding mechanism that enables lightweight VLMs to estimate navigation progress through image-level comparative reasoning instead of textual history modeling.

The navigation state at timestep $t$ is represented as:

\begin{equation}
M_t =
\left(
\mathrm{STTB}_t,
\mathrm{VB}_t
\right),
\end{equation}

where $\mathrm{STTB}_t$ denotes short-term trajectory observations and $\mathrm{VB}_t$ denotes the visual baseline.

The short-term trajectory observations (STTB) maintain recent observations within the current navigation rollout:

\begin{equation}
\mathrm{STTB}_t =
\mathrm{Update}\!\left(
\mathrm{STTB}_{t-1},
o_t^{0^\circ},
\mathbf{A}_t
\right),
\label{STTB_Update}
\end{equation}

where $o_t^{0^\circ}$ denotes the forward-view observation and $\mathbf{A}_t$ represents the executed action sequence. Rather than maintaining full trajectory history, STTB preserves only recent observations to reduce accumulated noisy or misleading historical information.

The visual baseline (VB) maintains an evolving visual reference for long-horizon state grounding:

\begin{equation}
\mathrm{VB}_t =
\mathrm{Update}\!\left(
\mathrm{VB}_{t-1},
b_t
\right),
\label{VB_Update}
\end{equation}
 
where $b_t$ denotes the current baseline observation. Instead of maintaining full trajectories or explicit maps, VB provides a stable visual reference for judging long-term navigation progress through image-level comparison.

Within this dual-reference formulation, STTB captures short-term trajectory dynamics, while VB provides a long-term visual reference for progress consistency checking across time. Both representations are jointly provided to the lightweight VLM together with the current observation, forming a multi-reference input space for inference.

Lightweight VLMs perform image-level comparison between current observations, STTB, and VB to infer progress direction without relying on textual summaries. This process avoids explicit maps and geometric state estimation, and instead relies on relative visual consistency across observations. Thus, the VLM derives progress and positional information implicitly during image‑based comparison, without constructing explicit spatial states.

\subsection{Stage-wise Execution–Transition Decoupling (SETD)}
\label{sec:setd}

Long-horizon VLN-CE requires lightweight VLMs to simultaneously handle sub-goal identification, local action execution, and stage transition within a unified inference process. Such entangled decision-making across multiple stages substantially increases reasoning complexity, exceeds the limited capacity of lightweight VLMs, and leads to error accumulation in long-horizon navigation.

To address this issue, we propose the \textbf{Stage-wise Execution--Transition Decoupling (SETD)} paradigm, which alleviates task entanglement by decomposing navigation into ordered sub-goals and decoupling local action execution from global stage transition, allowing each inference step to focus on a single decision objective.

An instruction $x$ is first decomposed into
\begin{equation}
G = [g_1, g_2, \dots, g_K],
\label{eq:sep_decompose}
\end{equation}
where each $g_k$ is a progressively executable and explicitly verifiable subgoal.

\paragraph{Execution stage}
The execution stage focuses exclusively on short-horizon action reasoning under the current sub-goal, avoiding interference from stage transition decisions. Local inference is performed using forward-centered observations
\begin{equation}
O_t^{\mathrm{loc}} =
\left\{o_t^{-30^\circ},\, o_t^{0^\circ},\, o_t^{+30^\circ}\right\},
\label{eq:local_obs}
\end{equation}
where $O_t^{\mathrm{loc}}$ denotes the local observation at step $t$, composed of three forward-centered views ($-30^\circ$, $0^\circ$, and $+30^\circ$).

Execution-stage evidence is then filtered from local observations together with STTB and VB:
\begin{equation}
\mathbf{e}_t^{\mathrm{execution}} =
\mathrm{Filter}_{\mathrm{execution}}\!\left(
g_{k_t},
O_t^{\mathrm{loc}},
\mathrm{STTB}_t,
\mathrm{VB}_t
\right),
\label{eq:execution_filter}
\end{equation}
where DGMF extracts sub-goal-relevant evidence from local observations together with grounded state references. STTB provides short-term motion consistency within the current rollout, while VB maintains long-range visual grounding across stages.

Based on the filtered evidence, the execution policy focuses on a single action objective and predicts a short-horizon action sequence:
\begin{equation}
\mathbf{A}_t \sim \pi_{\mathrm{execution}}\!\left(
g_{k_t},
\mathbf{e}_t^{\mathrm{execution}},
\mathrm{STTB}_t,
\mathrm{VB}_t
\right).
\label{eq:execution_policy}
\end{equation}

After each rollout, recent observations and executed actions are summarized as $\Sigma_t$ for transition-stage verification. On the final sub-goal, the execution stage may emit an early-stop signal when terminal evidence becomes sufficiently strong. This signal only terminates the current rollout, while the final completion decision is still handled by the transition stage.

\paragraph{Transition stage}
The transition stage performs stage-level verification to determine whether the current sub-goal should continue or switch to the next one. To support broader environmental awareness, transition reasoning uses panoramic observations
\begin{equation}
O_t^{\mathrm{pan}} =
\left\{o_t^{0^\circ},\, o_t^{60^\circ},\, o_t^{120^\circ},\, o_t^{180^\circ},\,
o_t^{240^\circ},\, o_t^{300^\circ}\right\},
\label{eq:pan_obs}
\end{equation}
which provide a global scene view for sub-goal verification.

Transition-stage evidence is filtered from panoramic observations together with VB and the rollout summary:
\begin{equation}
\mathbf{e}_t^{\mathrm{trans}} =
\mathrm{Filter}_{\mathrm{trans}}\!\left(
g_{k_t},
O_t^{\mathrm{pan}},
\mathrm{VB}_t,
\Sigma_t
\right),
\label{eq:transition_filter}
\end{equation}
where DGMF extracts transition-oriented evidence related to sub-goal completion, global progress consistency, and stage switching. Unlike execution, transition reasoning relies more heavily on VB and rollout summaries for stage-level progress verification.

Based on the filtered evidence, the transition policy predicts the stage-transition decision
\begin{equation}
\begin{aligned}
u_t &\sim \pi_{\mathrm{trans}}\!\left(
g_{k_t},
\mathbf{e}_t^{\mathrm{trans}},
\mathrm{VB}_t,
\Sigma_t
\right), \\
u_t &\in \{\mathrm{continue}, \mathrm{switch}\}.
\end{aligned}
\label{eq:transition}
\end{equation}
where $\pi_{\mathrm{trans}}$ denotes the transition-stage policy and $\mathbf{e}_t^{\mathrm{trans}}$ denotes the transition-stage evidence.

The sub-goal index is updated by
\begin{equation}
k_{t+1} = k_t + \mathbb{I}[u_t = \mathrm{switch}],
\label{eq:stage_update}
\end{equation}
where $\mathrm{switch}$ advances execution to the next sub-goal. For the final sub-goal, $\mathrm{switch}$ indicates task completion.

\section{Experiments}
\label{sec:experiments}

\begin{table*}[t]
\centering
\scriptsize
\setlength{\tabcolsep}{4pt}
\resizebox{\textwidth}{!}{
\begin{tabular}{l|c|c|c|c|ccccc}
\hline
Method & Backbone / Size & Map / Prior & WP & View & TL & NE$\downarrow$ & OSR$\uparrow$ & SR$\uparrow$ & SPL$\uparrow$ \\
\hline
MapGPT-CE~\cite{chen2024mapgpt} & GPT-4o ($\sim$200B) & Online map & No & Panoramic & 12.63 & 8.16 & 22.0 & 7.0 & 5.04 \\
DiscussNav~\cite{qiao2025open} & GPT-4 ($\sim$1.76T) & No & No & Panoramic & 6.27 & 7.77 & 15.0 & 11.0 & 10.51 \\
InstructNav~\cite{long2024instructnav} & GPT-4 ($\sim$1.76T) & Value maps & No & Egocentric & 7.74 & 6.89 & -- & 31.0 & 24.0 \\
Open-Nav~\cite{qiao2025open} & Llama3.1-70B & No & Yes & Panoramic & 8.07 & 7.25 & 23.0 & 16.0 & 12.90 \\
Open-Nav~\cite{qiao2025open} & GPT-4 ($\sim$1.76T) & No & Yes & Panoramic & 7.68 & 6.70 & 23.0 & 19.0 & 16.10 \\
CA-Nav~\cite{chen2025constraint} & GPT-4 ($\sim$1.76T) & Constraint prior & No & Egocentric & -- & 8.32 & 39.4 & 23.0 & 11.0 \\
SmartWay~\cite{shi2025smartway} & GPT-4o ($\sim$200B) & No & Yes & Panoramic & $13.09 $ & $7.01$ & $\mathbf{51.0}$ & $29.0 $ & $22.46$ \\
Fast-SmartWay~\cite{shi2025fast} & GPT-4o ($\sim$200B) & No & No & Frontal View & $12.56 \pm 0.71$ & $7.72 \pm 0.42$ & -- & $27.75 \pm 2.22$ & $\mathbf{24.95 \pm 2.70}$ \\
\hline
\textbf{LightZeroNav} & \textbf{Qwen3-VL-4B} & \textbf{No} & \textbf{No} & \textbf{Role-separated} & $13.63 \pm 0.67$ & $7.45 \pm 0.34$ & $39.0 \pm 6.0$ & $27.0 \pm 3.0$ & $18.0 \pm 2.0$ \\
\textbf{LightZeroNav} & \textbf{Qwen3-VL-8B} & \textbf{No} & \textbf{No} & \textbf{Role-separated} & $18.30 \pm 1.06$ & $8.32 \pm 0.07$ & $45.0 \pm 3.0$ & $30.0 \pm 2.0$ & $20.0 \pm 2.0$ \\
\textbf{LightZeroNav} & \textbf{Qwen3-VL-32B} & \textbf{No} & \textbf{No} & \textbf{Role-separated} & $19.19 \pm 1.88$ & $7.65 \pm 0.23$ & $\mathbf{51.0 \pm 4.0}$ & $\mathbf{38.0 \pm 1.0}$ & $22.0 \pm 1.0$ \\
\hline
\end{tabular}
}
\caption{Comparison with representative zero-shot VLN-CE methods.}
\label{tab:main_compare}
\vspace{2pt}
\begin{minipage}{\textwidth}
\footnotesize
\textit{Note:} All results in Table \ref{tab:main_compare}, \ref{tab:ablation_modules}, \ref{tab:ablation_SETD} are averaged over three runs and reported as mean $\pm$ standard deviation. Map / Prior indicates whether a method relies on explicit maps or additional spatial priors; WP denotes waypoint predictor; View denotes the sensing regime. Role-separated Dual FOV denotes execution stage relies on frontal views, while transition stage uses panoramic views; The DiscussNav results are reported following \cite{qiao2025open}, while the MapGPT-CE results follow \cite{shi2025smartway}. Approximate sizes for GPT-4 ($\sim$1.76T) and GPT-4o ($\sim$200B) follow the estimates summarized in \cite{abacha2025medec}.
\end{minipage}
\end{table*}

\begin{table*}[t]
\centering
\footnotesize
\renewcommand{\arraystretch}{1.0}
\setlength{\tabcolsep}{1.5pt}
\begin{tabular}{l|ccc|cccc|cccc}
\hline
Variant & Dual FOV & DGMF & ISGR& SR$\uparrow$ & OSR$\uparrow$ & SPL$\uparrow$ & nDTW$\uparrow$ & Steps$\downarrow$ & NE$\downarrow$ & TL$\downarrow$ & Coll.$\downarrow$ \\
\hline
Full & \checkmark & \checkmark & \checkmark & $\mathbf{30.0 \pm 2.0}$ & $45.0 \pm 3.0$ & $\mathbf{20.0 \pm 2.0}$ & $\mathbf{18.0 \pm 1.0}$ & $\mathbf{138.8 \pm 6.4}$ & $\mathbf{8.32 \pm 0.07}$ & $\mathbf{18.30 \pm 1.06}$ & $0.27 \pm 0.00$ \\
w/o DGMF & \checkmark &  & \checkmark & $12.0 \pm 6.0$ & $33.0 \pm 3.0$ & $6.0 \pm 2.0$ & $8.0 \pm 2.0$ & $170.1 \pm 3.2$ & $10.19 \pm 1.03$ & $23.46 \pm 1.27$ & $\mathbf{0.24 \pm 0.01}$ \\
w/o ISGR& \checkmark & \checkmark &  & $17.0 \pm 2.0$ & $52.0 \pm 2.0$ & $6.0 \pm 1.0$ & $4.0 \pm 1.0$ & $220.9 \pm 6.3$ & $10.37 \pm 0.70$ & $31.39 \pm 1.31$ & $0.30 \pm 0.01$ \\
Dual FOV Only & \checkmark &  &  & $14.0 \pm 3.0$ & $35.0 \pm 2.0$ & $7.0 \pm 1.0$ & $6.0 \pm 1.0$ & $189.6 \pm 2.8$ & $11.48 \pm 0.64$ & $25.37 \pm 0.52$ & $0.27 \pm 0.01$ \\
w/o Dual FOV &  & \checkmark & \checkmark & $23.0 \pm 5.0$ & $\mathbf{55.0 \pm 2.0}$ & $8.0 \pm 2.0$ & $4.0 \pm 1.0$ & $217.1 \pm 5.0$ & $9.23 \pm 0.39$ & $27.94 \pm 0.90$ & $0.33 \pm 0.01$ \\
\hline
\end{tabular}
\caption{Module ablation of LightZeroNav with the 8B backbone.}
\label{tab:ablation_modules}
\vspace{2pt}
\begin{minipage}{\textwidth}
\footnotesize
\textit{Note:} ISGR denotes Image-level State Grounding Representation. SR, OSR, SPL, and nDTW are reported in percentage points.
\end{minipage}
\end{table*}

\begin{table*}[t]
\centering
\footnotesize
\renewcommand{\arraystretch}{1.1}
\setlength{\tabcolsep}{3pt}
\begin{tabular}{l|ccc|cccc|cccc}
\hline
Variant & Dec. & Exe. & Trans. & SR$\uparrow$ & OSR$\uparrow$ & SPL$\uparrow$ & nDTW$\uparrow$ & Steps$\downarrow$ & NE$\downarrow$ & TL$\downarrow$ & Coll.$\downarrow$ \\
\hline
Full & \checkmark & \checkmark & \checkmark & $\mathbf{30.0 \pm 2.0}$ & $\mathbf{45.0 \pm 3.0}$ & $\mathbf{20.0 \pm 2.0}$ & $18.0 \pm 1.0$ & $138.8 \pm 6.4$ & $\mathbf{8.32 \pm 0.07}$ & $18.30 \pm 1.06$ & $0.27 \pm 0.00$ \\
w/o Transition & \checkmark & \checkmark &  & $11.0 \pm 3.0$ & $21.0 \pm 3.0$ & $10.0 \pm 2.0$ & $34.0 \pm 1.0$ & $43.8 \pm 6.6$ & $8.53 \pm 0.06$ & $7.28 \pm 1.17$ & $0.18 \pm 0.01$ \\
Execution Only &  & \checkmark &  & $8.0 \pm 3.0$ & $10.0 \pm 3.0$ & $7.0 \pm 3.0$ & $\mathbf{36.0 \pm 1.0}$ & $\mathbf{24.3 \pm 4.0}$ & $8.42 \pm 0.41$ & $\mathbf{3.83 \pm 0.54}$ & $\mathbf{0.16 \pm 0.01}$ \\
\hline
\end{tabular}
\caption{Ablation of the staged execution structure with the 8B backbone.}
\label{tab:ablation_SETD}
\vspace{2pt}
\begin{minipage}{\textwidth}
\footnotesize
\textit{Note:} Dec., Exe., and Trans. denote Decomposition, Execution, and Transition, respectively.
\end{minipage}
\end{table*}

\begin{figure*}[t]
    \centering
    \includegraphics[width=1\linewidth]{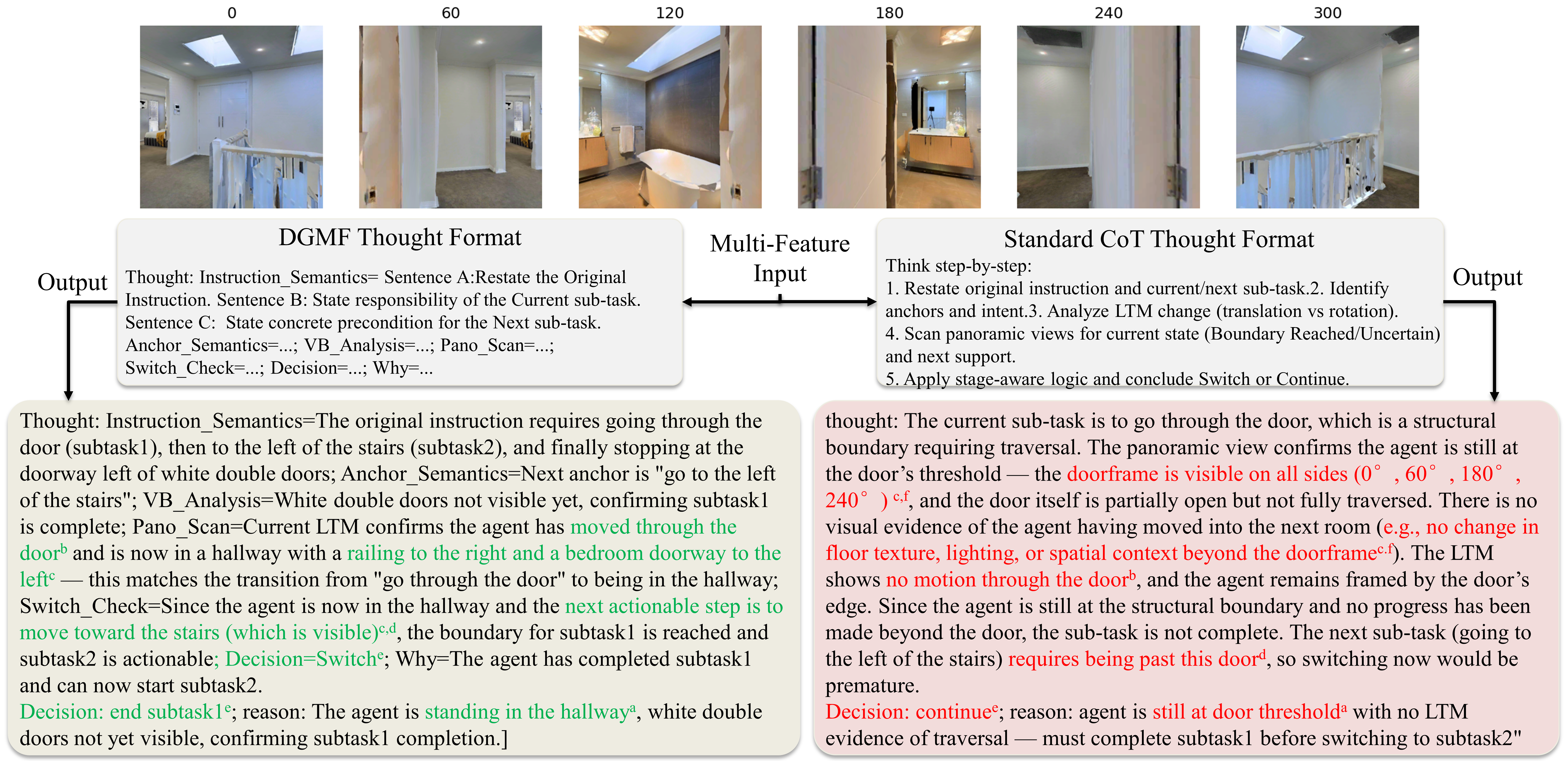}
\caption{
Qualitative comparison of reasoning traces between LightZeroNav and standard CoT in VLN-CE. The grey panels (left) illustrate the structured reasoning of LightZeroNav, where green text highlights its accurate identification of spatial transitions and decision-relevant anchors. In contrast, the pink panels (right) depict the reasoning path of standard CoT under the same scenario, where red text denotes inaccurate grounding of environmental boundaries and erroneous progress estimations. Superscripts denote specific reasoning functions: (a) static self-location, (b) dynamic progress grounding via ISGR, (c) decision-relevant evidence extraction, (d) goal-aligned task anticipation, (e) execution outcome, and (f) reasoning explosion because of information redundancy. The comparison demonstrates how structured framework of LightZeroNav mitigates the logical drift common in standard CoT.
}
    \label{fig:5}
\end{figure*}

\subsection{Experimental Setup}

We evaluate LightZeroNav on the Habitat-based VLN-CE benchmark, following the standard 100-episode evaluation protocol commonly used in prior zero-shot VLN-CE work for fair comparison~\cite{qiao2025open}. All methods are evaluated in a zero-shot setting without task-specific fine-tuning. We instantiate LightZeroNav as a multi-level enhancement framework for lightweight VLMs, where Stage-wise Execution–Transition Decoupling (SETD), Image-level State Grounding Representation (ISGR), and Decision-guided Multi-Feature Filtering (DGMF) are respectively designed to alleviate task entanglement, progress estimation inaccuracy, and information redundancy via stage-specific structured prompts over a shared Qwen3-VL backbone. The same framework is applied consistently across the 4B, 8B, and 32B variants.

We use standard VLN-CE metrics, including trajectory length (TL), navigation error (NE), oracle success rate (OSR), success rate (SR), and success weighted by path length (SPL). For ablation studies, we additionally report nDTW, number of steps, and collision rate to better characterize execution behavior.

\subsection{Comparison with zero-shot VLN-CE Baselines}

The comparison is challenging because several strong baselines rely on RGB-D input, waypoint prediction, or additional spatial priors, whereas our method uses RGB only without waypoint supervision or environment-specific training. We take SmartWay as the main reference because it is the strongest reported baseline in our comparison set.

We observe that LightZeroNav-32B raises SR from 29.0 to 38.0 while matching the best OSR at 51.0 as shown in Table~\ref{tab:main_compare}. This is the clearest evidence that the gain does not come simply from passing near the goal more often, but from converting goal vicinity into successful stopping more reliably. The 8B variant also remains competitive, reaching 30.0 SR and 45.0 OSR, while the 4B variant attains 27.0 SR. The trend across 4B, 8B, and 32B indicates that the proposed multi-level inference framework is not tied to a single backbone scale, but provides a consistent benefit across model capacities. This is particularly relevant for lightweight backbones, indicating that the gains come not only from stronger raw model capacity, but also from alleviating task entanglement through SETD, reducing information redundancy through DGMF, and improving progress estimation through ISGR. It also helps explain why the 8B model obtains competitive performance with GPT-4o($\sim$200B). These results suggest that alleviating task entanglement, reducing information redundancy, and im-
proving progress grounding can be comparable in impact to
increasing model scale in zero-shot VLN-CE.

Compared with methods that rely on broader perception\cite{shi2025smartway,wang2025dreamnav}, waypoint prediction\cite{qiao2025open,shi2025smartway}, or external spatial scaffolds\cite{chen2025constraint}, the LightZeroNav obtains its gains mainly on completion-oriented metrics rather than on shortest-path behavior alone. In other words, its advantage is less about aggressively reducing NE or TL at every step, and more about making long-horizon execution easier to sustain and finish correctly. This is consistent with the design of SETD, ISGR, and DGMF, which together reduce stage confusion, weaken drift accumulation, and help the model stop at the right time instead of only moving toward the goal region. 

\subsection{Ablation Study}

We conduct ablation study on the 8B backbone to examine how each design contributes to the final result.

The clearest drop appears when DGMF is removed. SR falls from 30.0 to 12.0, SPL drops from 20.0 to 6.0, as listed in Table \ref{tab:ablation_modules}, and the agent takes longer and less efficient trajectories. This confirms that under limited model capacity, information redundancy is the dominant bottleneck, and highlights the effectiveness of Decision-guided Multi-Feature Filtering (DGMF) in improving decision signal quality. Once the model must reason over unfiltered visual context, decisions become less focused and long-horizon execution degrades quickly. This is consistent with the weakness of lightweight models under redundancy-heavy input, where limited capacity makes implicit evidence selection less reliable.

Removing ISGR produces a different failure pattern. SR decreases from 30.0 to 17.0 and SPL from 20.0 to 6.0, while steps, TL, NE, and collisions all increase noticeably. Meanwhile, OSR increases from 45.0 to 52.0, suggesting that the agent more frequently reaches the goal vicinity. However, the decrease in SR indicates that such proximity is not consistently converted into successful task completion. Without the comparison between STTB and VB, the agent becomes prone to progress misestimation caused by weakened progress grounding consistency. It reaches the goal vicinity more frequently but fails to reliably convert perceived proximity into correct termination decisions, demonstrating the importance of ISGR for progress grounding. In fact, ISGR provides lightweight VLMs with stable progress cues, reducing information loss, hallucination accumulation, and semantic drift in long-horizon navigation by relying on image-level comparative reasoning instead of textual history modeling.

The ablation of the dual FOV design shows another distinct effect. Removing it lowers SR from 30.0 to 23.0 and SPL from 20.0 to 8.0, while OSR increases to 55.0. This indicates that insufficient functional decoupling between perception streams degrades stage transition accuracy, leading to inconsistency between goal detection and switching decisions. Using broader or less role-specific perception everywhere may help the agent notice more candidate cues, but it also makes local rollout less stable and weakens the separation between action generation and switching decisions.

Table~\ref{tab:ablation_SETD} further shows that the SETD is critical. Removing the transition stage causes SR to drop from 30.0 to 11.0 and OSR from 45.0 to 21.0, even though trajectories become shorter. This shows that execution-only inference is insufficient for long-horizon decision making without explicit stage decoupling. Without an explicit stage-level decision process, the agent may move, but it cannot reliably decide when the current subgoal is complete or when to switch. When both task decomposition and transition are removed, the system reduces to a single-stage execution process. In this setting, SR and OSR drop even further, while nDTW becomes higher. This apparent inconsistency is expected. Shorter and earlier-terminated trajectories can improve path-shape similarity metrics without improving actual task completion. Here, higher nDTW reflects conservative or truncated motion rather than successful long-horizon navigation.

Figure~\ref{fig:5} provides a qualitative interpretation of the ablation results using the lightweight Qwen3-VL-8B VLM backbone.
Standard CoT reasoning entangles self-location estimation, progress grounding, sub-goal anticipation, and transition verification within a single inference process, leading to reasoning-space explosion and unstable stage switching, as highlighted by annotation $(f)$.
In contrast, LightZeroNav decomposes these reasoning functions into structured stages. 
In fact, the decoupled execution–transition mechanism of the SETD reduces task entanglement, mitigates cross-stage interference, and alleviates long-horizon error accumulation for lightweight VLMs. Annotations $(a)$ and $(b)$ illustrate how ISGR supports static self-location judgment and dynamic progress grounding through comparison between the current observation, STTB, and VB. 
Annotations $(c)$ and $(d)$ show how DGMF suppresses redundant multi-view observations and preserves subgoal-relevant evidence for the current decision stage. 
Annotation $(e)$ demonstrates more stable execution and transition behavior under structured reasoning.
These qualitative observations are consistent with Tables~\ref{tab:ablation_modules} and~\ref{tab:ablation_SETD}: removing DGMF weakens decision quality under redundancy-heavy inputs, removing the ISGR degrades progress grounding consistency, and removing the transition stage destabilizes long-horizon execution.

Overall, the ablation study directly validates the proposed three-level bottleneck decomposition in lightweight VLM-based VLN-CE. The DGMF addresses information redundancy and attention dilution, The ISGR mitigates progress estimation drift through progress grounding consistency, and the SETD resolves task entanglement through explicit execution–transition decoupling. Their combination converts overloaded inference into structured, low-burden decision making for long-horizon navigation.

\section{Conclusion}

We present LightZeroNav, a lightweight zero-shot framework for vision language navigation in continuous environments (VLN-CE), as a multi-level enhancement framework for lightweight VLMs. Instead of treating VLN-CE as a direct observation-to-action mapping problem, the LightZeroNav formulates it as stage-wise execution over ordered subgoals with explicit execution–transition decoupling (SETD), combines Image-level State Grounding Representation (ISGR), and uses Decision-guided Multi-Feature Filtering (DGMF) to support both local rollout and stage switching. 

Extensive tests on VLN-CE show that the LightZeroNav improves zero-shot continuous navigation with open-source VLMs without task-specific training, or waypoint supervision, demonstrating its effectiveness in zero-shot VLN-CE. Future work will focus on improving subgoal transition accuracy and failure recovery efficiency while preserving the simplicity and generality of the zero-shot setting.




\bibliographystyle{IEEEtran}
\bibliography{refs}


 




\vfill

\end{document}